\documentclass{article}

\PassOptionsToPackage{square}{natbib}
\PassOptionsToPackage{sort&compress}{natbib}
\PassOptionsToPackage{comma}{natbib}
\usepackage[preprint]{neurips_2024}
\usepackage{verbatim}

\usepackage[utf8]{inputenc} 
\usepackage[T1]{fontenc}    
\usepackage{url}            
\usepackage{booktabs}       
\usepackage{amsfonts}       
\usepackage{nicefrac}       
\usepackage{microtype}      
\usepackage{xcolor}         

\usepackage{adjustbox}
\usepackage{natbib}
\setcitestyle{numbers,square}
\usepackage{multirow} 
\usepackage{bm}
\usepackage{amsmath}
\usepackage{pifont}
\newcommand{\cmark}{\ding{51}}%
\newcommand{\xmark}{\ding{55}}%
\newcommand{\ie}{\textit{i.e.}}
\newcommand{\eg}{\textit{e.g.}}
\usepackage{color, colortbl}
\definecolor{citecolor}{HTML}{2980b9}
\definecolor{linkcolor}{HTML}{c0392b}

\usepackage[hidelinks,breaklinks=true,colorlinks,bookmarks=false,citecolor=citecolor,linkcolor=linkcolor]{hyperref}

\title{VividPose: Advancing Stable Video Diffusion for Realistic Human Image Animation}

\author{%
   Qilin Wang$^{1}$\thanks{Equal contribution.}
  ~~Zhengkai Jiang$^{2*}$
  ~~Chengming Xu$^2$
  ~~Jiangning Zhang$^{2}$ \\ 
  ~~\bf{Yabiao Wang}$^2$ 
  ~~\bf{Xinyi Zhang}$^{2}$ 
  ~~\bf{Yun Cao}$^{2}$
  ~~\bf{Weijian Cao}$^2$
  ~~\bf{Chengjie Wang}$^2$
  ~~\bf{Yanwei Fu}$^{1}$\thanks{Corresponding Author} \\
  \textsuperscript{1}Fudan University, Shanghai, China~~~
  \textsuperscript{2}Tencent Youtu Lab, Shanghai, China~~~\\
  \texttt{\url{https://Kelu007.github.io/vivid-pose/}}
}

\begin{document}

\maketitle

\begin{abstract}
Human image animation involves generating a video from a static image by following a specified pose sequence. 
Current approaches typically adopt a multi-stage pipeline that separately learns appearance and motion, which often leads to appearance degradation and temporal inconsistencies. 
To address these issues, we propose VividPose, an innovative end-to-end pipeline based on Stable Video Diffusion (SVD) that ensures superior temporal stability. 
To enhance the retention of human identity, we propose an identity-aware appearance controller that integrates additional facial information without compromising other appearance details such as clothing texture and background. 
This approach ensures that the generated videos maintain high fidelity to the identity of human subject, preserving key facial features across various poses. 
To accommodate diverse human body shapes and hand movements, we introduce a geometry-aware pose controller that utilizes both dense rendering maps from SMPL-X and sparse skeleton maps. 
This enables accurate alignment of pose and shape in the generated videos, providing a robust framework capable of handling a wide range of body shapes and dynamic hand movements. 
Extensive qualitative and quantitative experiments on the UBCFashion and TikTok benchmarks demonstrate that our method achieves state-of-the-art performance. 
Furthermore, VividPose exhibits superior generalization capabilities on our proposed in-the-wild dataset.
Codes and models will be available. 
\end{abstract}
\section{Introduction}


Human image animation involves generating dynamic and realistic videos from static human images by following a given sequence of poses. 
This technology has garnered significant attention due to its wide range of applications across various domains. In social media, it can be used to create engaging and personalized content. In the movie and entertainment industry, it facilitates the production of lifelike character animations. In online retail, it enhances virtual try-ons by animating mannequins. The ability to animate static images into vivid and coherent videos has immense value, making it a highly explored area in both academic research and practical applications.



With the advent of powerful generative models like Generative Adversarial Networks (GANs) and Diffusion Models (DMs), human image animation has advanced significantly. Recent DM-based methods~\cite{DreamPose,DISCO,FollowYourPose,MagicAnimate,AnimateAnyone,PIDM,PoseAnimate,MagicPose}  generate animated videos by combining appearance information from a reference image with pose sequences from a driving video. Particularly, 
by fully utilizing the prior knowledge learned by DMs from large-scale datasets, these methods significantly improve  generation quality, avoiding issues commonly seen in GAN-based methods~\cite{FOMM,TianRCO0MT21,G3AN} such as local deformation artifacts, insufficient appearance details and temporal inconsistencies. Specifically, methods like AnimateAnyone~\cite{AnimateAnyone}, MagicAnimate~\cite{MagicAnimate}, and MagicPose~\cite{MagicPose} utilize ReferenceNet (a variant of  denoising UNet~\cite{UNet}) to encode multi-scale appearance information, effectively preserving fine-grained details such as background and clothing textures. However, these methods model the appearance information of the entire human image as a whole, neglecting the domain gap between facial features and other details like clothing. This often leads to models focusing excessively on clothing textures and background, while failing to maintain high face identity consistency. 

Moreover, since pretrained SD is unaware of motion priors, these approaches mainly rely on extra temporal layers to incorporte such knowledge into the diffusion process. However, compared to the the original parameters  trained on extensive datasets, these new parameters are only exposed to limited data due to difficulty of collecting high-quality human animating videos. This results in insufficiently learned motion knowledge, leading to issues like inter-frame jitter and temporal inconsistencies in the generated videos.
Additionally, the pose conditions used by these methods are derived from 2D estimators, which prevents the modification of pose sequences to align with the reference image's body shape. This misalignment results in inaccurate human body shape generation.
%
%

\begin{figure*}[t!]
  \centering
\includegraphics[width=\textwidth]{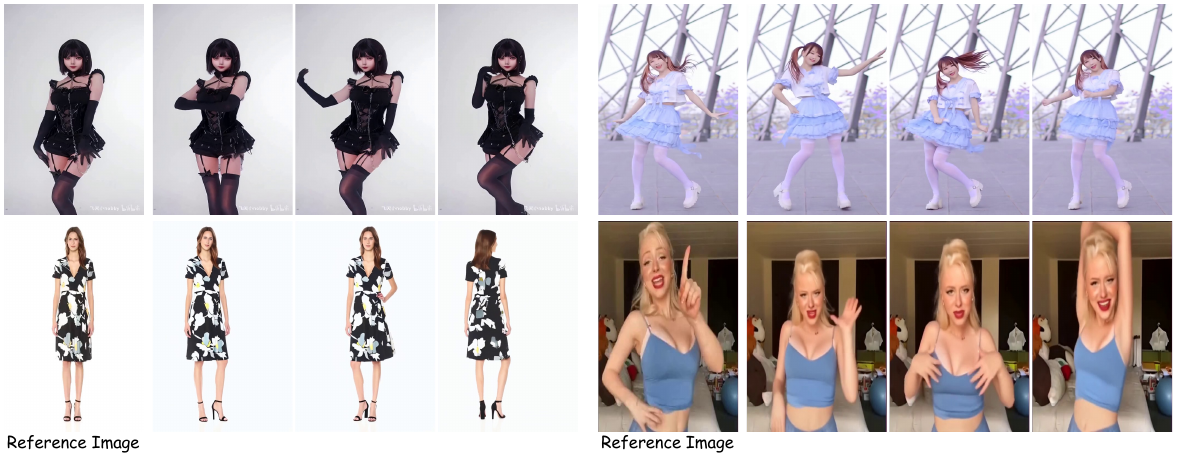}
   \caption{\textbf{Temporally coherent and consistent animation results generated by VividPose}, given a reference image (the leftmost image in each group). Our method excels at producing smooth and stable videos, even for large movements and complex hand gestures, while preserving the appearance details of the reference image. More video results are available in the supplementary material.} 
    \label{fig_teaser}
\end{figure*}

To address these limitations, we propose VividPose, a novel human image animation framework driven by the powerful motion-aware prior knowledge learned by Stable Video Diffusion (SVD). 
Ideally, such prior knowledge can significantly enhance the temporal consistency and smoothness of human animation videos compared with current SD based methods. To adapt SVD for human animation, VividPose presents an identity-aware appearance controller to improve face identity retention and a geometry-aware pose controller to accommodate diverse human body shapes and hand movements.
%
%
%
These innovations ensure that our approach maintains high visual fidelity and effectively handles a wide range of human poses and appearances as shown in Fig.~\ref{fig_teaser}.


In specific, to enhance human identity retention of pretrained SVD, we introduce an identity-aware appearance controller that incorporates additional facial information without compromising other appearance details such as clothing texture and background details. We utilize ReferenceNet to encode multi-scale features from the reference image, which are then injected into the denoising UNet via spatial self-attention. High-level features are encoded using CLIP~\cite{CLIP}, while face identity features are encoded using ArcFace~\cite{ArcFace}; both are then injected into the UNet through a decoupled cross-attention mechanism. 
%
%
This controller ensures that the generated videos maintain high fidelity to  the human subject's identity, preserving key facial features across different poses. To accommodate diverse human body shapes and hand effects, we propose a geometry-aware pose controller. This controller leverages both dense rendering maps from a 3D parametric human model (\ie, SMPL-X~\cite{SMPLX}) and sparse skeleton maps extracted from a 2D pose estimator (\ie, DWPose~\cite{DWPose}). By establishing geometric correspondence between the reconstructed SMPL-X from the reference image and the SMPL-X-based pose sequences from the driving video, our approach ensures accurate body shape generation in the resulting videos. We use two lightweight encoders to separately encode features from the rendering maps and the skeleton maps, which are then fused through summation
%
%
To avoid disrupting SVD's inherent generative capabilities during initial training, we use zero convolutions at the end of each encoder. The fused feature is then added to the noise latent before being input into the UNet. This robust framework effectively handles a wide variety of body shapes and dynamic hand movements, ensuring high-quality and consistent animations.

Extensive qualitative and quantitative experiments on the UBCFashion and TikTok benchmarks demonstrate that VividPose achieves state-of-the-art results in terms of \textit{better temporal consistency}, \textit{higher visual fidelity} and \textit{stronger generalization ablility}. Additionally, VividPose exhibits superior performance on our proposed in-the-wild dataset, showcasing its robustness and effectiveness in real-world scenarios. 
In summary, we make several contributions:
   (1) We propose VividPose, a novel end-to-end pipeline leveraging Stable Video Diffusion, which significantly enhances the temporal consistency and smoothness of human animation videos. 
   (2) We introduce an identity-aware appearance controller that integrates additional facial information while preserving other appearance details such as clothing texture. This approach ensures high fidelity to the identity of the human subject across different poses.
   (3) We propose a geometry-aware pose controller that leverages both dense rendering maps and sparse skeleton maps, ensuring accurate body shape generation and effectively accommodating a wide range of body shapes and dynamic hand movements.

\section{Related Work}

\textbf{Diffusion Models for Video Generation.}
The success of diffusion models in image generation has inspired extensive research into their application for video generation. Early efforts in video generation~\cite{EsserCAGG23,VDM,CogVideo,Text2Video-Zero,FateZero,Make-A-Video,motionmaster,Tune-A-Video} have focused on extending image-based diffusion models with temporal components to address the dynamic nature of video sequences. For instance, Video LDM~\cite{VideoLDM} begins with a pretraining phase on static images, subsequently incorporating temporal layers that are finetuned on video data. AnimateDiff~\cite{AnimateDiff} offers a motion-specific module, pretrained on extensive video corpora, which can be integrated into pre-existing text-to-image architectures without additional modifications. Several studies have proposed incorporating inter-frame attention mechanisms. Methods like VideoCrafter~\cite{VideoCrafter} combine textual and visual features from CLIP~\cite{CLIP} as inputs for cross-attention layers to improve video generation quality. VideoComposer~\cite{VideoComposer} integrates images as conditional prompts during training, using them to guide the video generation process. Despite these advancements, the field continues to grapple with the challenge of producing stable, high-quality video outputs. The effective incorporation of static image conditions into video synthesis remains an active area of research. Our work builds upon these foundations, aiming to address these issues by leveraging Stable Video Diffusion~\cite{SVD} (SVD), which significantly enhances temporal coherence and the handling of complex motion dynamics in video generation.

\textbf{Human Image Animation.}
Human image animation involves creating a video by animating a static image based on given pose sequences.
GAN-based methods~\cite{FOMM,TianRCO0MT21,G3AN,RenLLL20,SiarohinLT0S19,MRAA,xu2022designing,YuPCZXL23,ZhangYLX22,TPS,ChanGZE19} typically first predict a flow map from the motion changes in the driving video. This motion representation is then used to spatially warp the reference human image, producing transformed frames. However, these methods often struggle to handle human self-occlusion and large motion variations during pose transfer, leading to insufficient appearance details and temporal inconsistencies. Moreover, GAN-based methods generally fail to produce satisfactory results when the identity of the reference image and the driving video differs. Recently, Diffusion Model (DM)-based methods~\cite{DreamPose,DISCO,FollowYourPose,MagicAnimate,AnimateAnyone,PIDM,PoseAnimate,MagicPose} have emerged as a more effective alternative for generating high-quality human animation videos. For instance, DreamPose~\cite{DreamPose} designs an adapter to integrate features encoded by VAE~\cite{VAE} and CLIP from the reference image. However, it requires test-time fine-tuning for specific images during inference, which is inefficient. DISCO~\cite{DISCO} addresses the test-time fine-tuning issue by using CLIP to encode foreground features and ControlNet to integrate background features. Nevertheless, it still struggles to retain fine-grained appearance details. Recent methods like AnimateAnyone~\cite{AnimateAnyone}, MagicAnimate~\cite{MagicAnimate}, and MagicPose~\cite{MagicPose} propose using ReferenceNet to encode fine-grained features and inject them via spatial self-attention. This approach enhances clothing textures and background details, but still falls short in maintaining face identity consistency. Additionally, these methods use skeleton maps from 2D estimators as pose conditions, which often leads to misalignment between the pose condition and the reference image. Finally, they typically add temporal layers to existing image generation architectures and perform multi-stage training, resulting in noticeable jitter and temporal inconsistency issues.
\section{Method}

\subsection{Overview}
\textbf{Preliminary: Stable Video Diffusion.}
SVD~\cite{SVD} is a cutting-edge video generation model that extends latent diffusion models from 2D image synthesis to high-resolution, temporally consistent video creation by taking text and image as inputs.  Technically, SVD introduces 3D convolution and temporal attention layers. The temporal layers are also integrated into the VAE decoder. A major improvement in SVD is the transition from the DDPM~\cite{DDPM} noise scheduler to the EDM~\cite{EDM} scheduler, which uses continuous noise scales \(\sigma_t\) for more flexible and effective sampling, replacing discrete timesteps. This end-to-end training paradigm and pipeline of SVD maintain strong temporal consistency in video generation, making it particularly suitable for human image animation. More details are presented in  Sec.~\ref{appendix:svd} of Appendix.

\textbf{Preliminary: SMPL-X.}  
It~\cite{SMPLX} is an advanced 3D parametric human model that enhances the capabilities of the SMPL~\cite{SMPL} model by incorporating detailed facial, hand, and body features. It represents the human body as a mesh with a fixed topology, controlled by shape, pose, and expression parameters.
Shape parameters (\(\beta\)) capture identity-specific attributes, pose parameters (\(\theta\)) encode joint rotations, and expression parameters (\(\psi\)) model facial expressions.
To generate a 2D rendering map \(I_{render}\) from the SMPL-X model, we employ a differentiable rendering process. Using the camera projection function (\(\Pi\)), the rendering process is formulated as:
\[
I_{render} = \Pi(M(\beta, \theta, \psi))
\]
Here, \(M(\beta, \theta, \psi)\) represents the deformed 3D human model. More details in Sec.~\ref{smplx} of Appendix.

\textbf{SMPL-X-base Pose Sequence.}  In our work, this process above accurately projects the 3D human model onto the 2D image plane. For human image animation, shape and expression parameters (\(\beta\) and \(\psi\)) are obtained from the reference image, while pose parameters (\(\theta\)) are extracted from the driving videos. The resulting rendering maps serve as pose conditions, ensuring that the generated video's body shape aligns accurately with the reference image while following the dynamic poses from the driving videos, resulting in realistic and coherent animations.



\begin{figure*}[t]
  \centering
\includegraphics[width=0.9\textwidth]{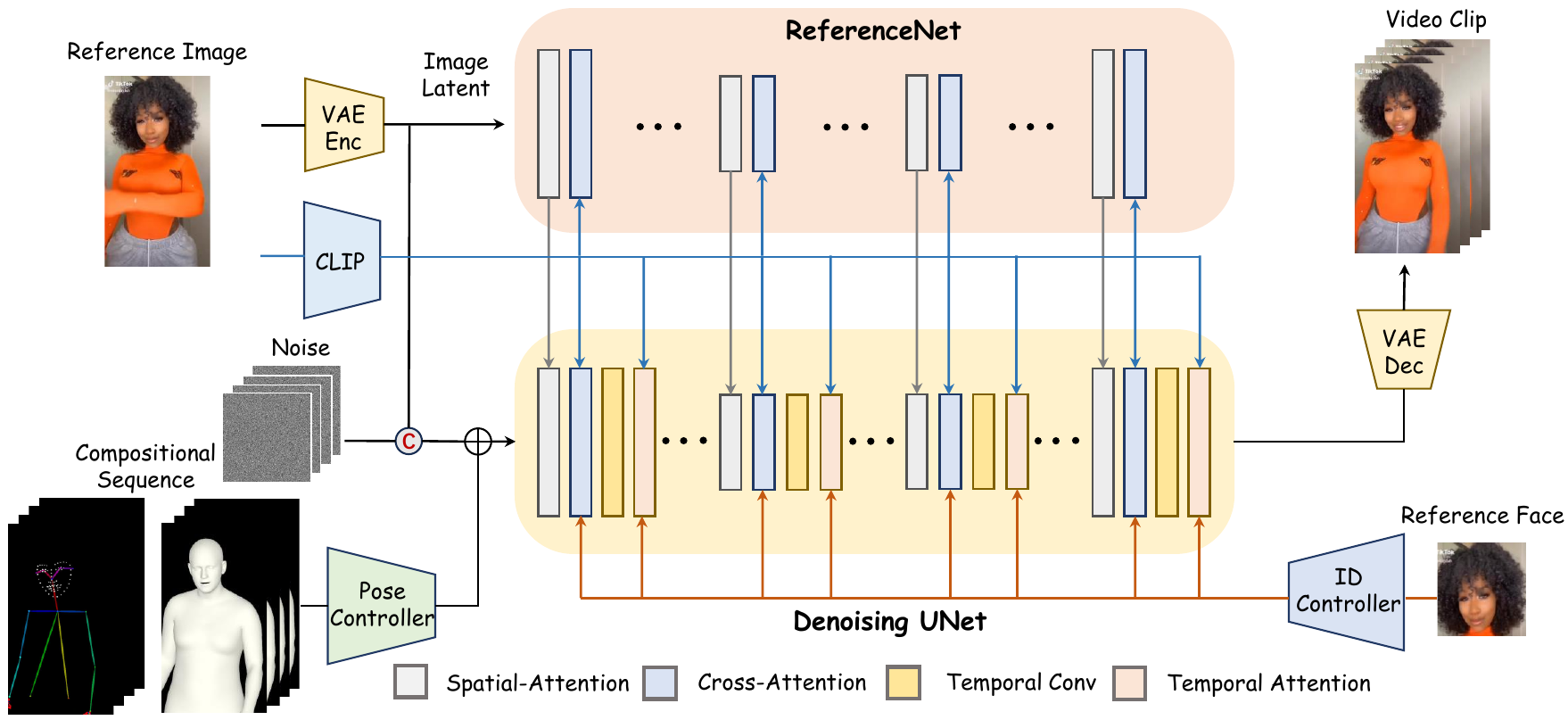}
   \caption{\textbf{Overview of VividPose}. The Denoising UNet (\ie, SVD), consists of Spatial-Attention, Cross-Attention, Temporal-Conv, and Temporal-Attention blocks. We utilize ReferenceNet to encode multi-scale features from the reference image, which are then injected into the denoising UNet via spatial self-attention. We also use CLIP to encode high-level features and ID Controller (\ie, ArcFace) to encode face identity features. These two features are injected via decoupled cross-attention. The compositional sequences consist of a skeleton map sequence extracted by DWPose and a rendering map sequence extracted by SMPLer-X. These sequences are initially encoded by the Pose Controller and then fused with noisy video frame latents. Furthermore, the reference image latent, generated by the VAE encoder, is concatenated with the noisy latents to preserve high-fidelity appearance details.} 
    \label{fig_framework}
\end{figure*}

\textbf{Overview}. Given a reference image \(I_{\text{ref}}\) and a pose sequence \(P^{1:F}\), where \(F\) is the number of frames, human image animation aims to generate a video \(I^{1:F}\) that maintains the appearance of \(I_{\text{ref}}\) while following the pose sequence \(P^{1:F}\). Current diffusion-based methods typically employ a multi-stage pipeline to separately learn appearance and motion, often resulting in appearance degradation and temporal inconsistency. To address these issues, we propose a novel end-to-end pipeline based on Stable Video Diffusion (SVD), as illustrated in Fig.~\ref{fig_framework}.

\subsection{Identity-aware Appearance Control}
As for human image animation, it is crucial to preserve the appearance information of \(I_{\text{ref}}\), such as face identity, clothing texture and background. Recent works typically use ReferenceNet (\ie, a copy of the denoising SD UNet) to encode low-level features of the whole reference human image. For each self-attention layer in the spatial block of SVD UNet, given the query features \(z_s\) and the reference feature \(f_r\), the output of self-attention \(z_s'\) can be defined as:
\begin{equation}
z_s' = \text{Attention}(Q_s, K_s, V_s) = \text{Softmax}\left(\frac{Q_s K_s^\top}{\sqrt{d}}\right) V_s
\end{equation}
where \(Q_s = z_s W_{qs}\), \(K_s = [z_s, f_r] W_{ks}\), and \(V_s = [z_s, f_r] W_{vs}\) are the query, key, and value matrices of the attention operation respectively. [$\cdot$] denotes concatenation operation. \(W_{qs}\), \(W_{ks}\), and \(W_{vs}\) are the weight matrices of the trainable linear projection layers. Moreover, these methods leverage the high-level reference features \(f_{clip}\) extracted via CLIP~\cite{CLIP} to adapt the cross-attention layers in the spatial block and temporal block of UNet. Specifically, given the query features \(z_c\) and the image CLIP feature \(f_{clip}\), the output of cross-attention \(z_c'\) can be defined as follows:
\begin{equation}
z_c' = \text{Attention}(Q_c, K_c, V_c) = \text{Softmax}\left(\frac{Q_c K_c^\top}{\sqrt{d}}\right) V_c
\end{equation}
where \(Q_c = z_c W_{qc}\), \(K_c = f_{clip} W_{kc}\), and \(V_c = f_{clip} W_{vc}\).

Although this approach effectively preserves clothing texture details and background, it often struggles to maintain human face identity due to the domain gap between faces and clothing textures. The ReferenceNet mainly focuses on low-level information, hence cannot well handle identity related features. The CLIP feature \(f_{clip}\) indeed provides high-level global features about the reference image, but in general human faces do not dominate the reference image, hence they would not be highlighted in \(f_{clip}\). To this end, we propose an identity-aware appearance controller to extract and compensate \(f_{clip}\) with identity-aware facial features. Specifically, we use ArcFace~\cite{ArcFace} to extract facial features \(f_{id}\) from \(I_{\text{ref}}\). Then \(f_{id}\) is injected into the UNet in the same way as \(f_{clip}\), \ie through decoupled cross-attention. In concrete, we map \(f_{id}\) through an MLP to the same dimension as \(f_{clip}\), obtaining \(f_{id'}\). We add a new cross-attention layer for each cross-attention layer in the original UNet to insert face identity feature. Given \(f_{id'}\), the output of the new cross-attention \(z_c''\) is computed as follows:
\begin{equation}
z_c'' = \text{Attention}(Q_c, K_c', V_c') = \text{Softmax}\left(\frac{Q_c (K_c')^\top}{\sqrt{d}}\right) V_c'
\label{eq:id-cross-attn}
\end{equation}
where \(Q_c = z_c W_{qc}'\), \(K_c' = f_{id} W_{kc}'\), and \(V_c' = f_{id} W_{vc}'\). Hence, the final formulation of the decoupled cross-attention is defined as follows:
\begin{equation}
z_c''' = z_c' + z_c''
\end{equation}

The additional attention as in Eq.~\ref{eq:id-cross-attn} enhances face identity without compromising other appearance information, such as clothing texture and background, thus ensuring the effective integration of fine-grained facial feature with the UNet features without disturbing other appearance features. 




\subsection{Geometry-aware Pose Control}
The human image animation pipeline takes a reference image and a pose sequence extracted from any driving video as input. Existing methods typically use either skeleton maps or dense maps (both derived from 2D pose estimator) as the pose driving signal. This often leads to shape misalignment issues. Specifically, skeleton map only captures pose information and ignores body shape details. While dense map includes both pose and shape information, the body shape is from the driving video and cannot be modified to align with the reference image's body shape. Consequently, this results in misalignment between the shape of the pose driving signal and the shape in the reference image, leading to inaccuracies in the generated video's human body shape.

To address this, we incorporate the rendering map from the 3D parametric human model SMPL-X as the pose driving signal. Benefiting from the parameterized representation of SMPL-X, we can establish geometric correspondence between the reconstructed SMPL-X from the reference image and the SMPL-X-based pose sequences extracted from the driving video. This alignment not only ensures the pose driving signal's shape matches the reference image's shape but also provides additional geometric relationships (\eg, hand crossing and body occlusions) from the 3D information. By extracting the shape and pose parameters from a reference image and a driving video respectively, we can create rendering maps that ensure the generated video's body shape aligns with the reference image while following the driving video's pose. This comprehensive representation is crucial for achieving realistic and coherent animations, capturing subtle nuances in hand movements, and handling variations in body shapes effectively.

Specifically, we utilize the 3D estimator SMPLer-X~\cite{SMPLerX} to estimate the shape, pose, and other coefficients from SMPL-X human model. The SMPL-X is then rendered to obtain 2D rendering maps as the pose driving signal. During inference, the shape parameters from the driving video are replaced with the shape parameters from the reference image, resulting in rendering maps that align the shape with the reference image while maintaining the pose information from the driving video.

We observe that the skeleton map provides some sparse pose constraints. Therefore, we combine the skeleton map \(I_{skeleton}\) extracted by DWPose~\cite{DWPose} with the SMPL-X rendering map \(I_{render}\), using them as sparse and geometry-aware dense conditions respectively. We use two lightweight encoders, \(\tau_1\) and \(\tau_2\) (each consisting of a set of convolutional layers and self-attention layers), to separately encode the rendering map and the skeleton map. To minimally disrupt the generative capabilities of SVD, we use zero convolution layers at the end of \(\tau_1\) and \(\tau_2\). The encoded features are then aggregated through summation to obtain the final fused pose condition \(f_{pose}\), as follows:

\begin{equation}
f_{pose} = \tau_1(I_{render}) + \tau_2(I_{skeleton})
\end{equation}

The $f_{pose}$ is added to the noise latent before being input into the UNet.

\section{Experiments}

\subsection{Settings}
\label{sec:settings}
\textbf{Datasets.} 
We conduct experiments on two commonly-used academic benchmarks, \ie, the UBCFashion~\cite{UBCFashion} and TikTok~\cite{TikTok} datasets. We also meticulously collect and process 3,000 videos from the internet to further validate our method's applicability in real-world scenarios. Specifically, UBCFashion comprises 500 training videos and 100 testing videos, each containing approximately 350 frames. This dataset is well-suited for fashion-related human animation tasks. TikTok consists of 340 single-person dance videos, each with a duration of 10-15 seconds. Most of TikTok videos focus on the upper body of human subject. We follow DISCO~\cite{DISCO} and use the same train and test split. Our curated dataset includes 2,224 dance videos from Bilibili and 776 videos from Douyin. These videos encompass a wide range of appearance and pose variations, such as indoor and outdoor scenes, diverse clothing textures, different age groups, and various dance styles. All frames from these datasets are extracted according to the original video's fps, and both DWPose~\cite{DWPose} and SMPLer-X~\cite{SMPLerX} are applied to each frame to infer the skeleton maps and rendering maps respectively. 

\textbf{Metrics.}
We assess both single-frame image quality and overall video fidelity to ensure a comprehensive evaluation. For single-frame quality, we use the L1 error, SSIM~\cite{SSIM}, LPIPS~\cite{LPIPS}, PSNR~\cite{PSNR} and FID~\cite{FID}. Video fidelity is evaluated through the FID-VID~\cite{FID-VID} and FVD~\cite{FVD}. These metrics allow us to rigorously measure the visual quality of individual frames and the temporal coherence of the generated videos, ensuring a thorough assessment of our method's performance.

\textbf{Implementation Details.}
During training, individual video frames are sampled, resized, and center-cropped to a resolution of 512$\times$768. We initialize the denoising UNet with SVD-img2vid~\cite{SVD} to generate 14 frames at a time and use SD2.1~\cite{SD} to initialize ReferenceNet. The denoising UNet, ReferenceNet, identity controller, and pose controller are trained in an end-to-end manner. All experiments are conducted on 8 NVIDIA 80GB H800 GPUs with a batch size of 32.

\textbf{Comparison Methods.}
Our evaluation includes a thorough comparison with a range of state-of-the-art methods in the domain of human image animation: (1) MRAA~\cite{MRAA} is a GAN-based method utilizing the optical flow estimations from driving sequences to warp the source image. Recent diffusion-based methods like (2) MagicAnimate~\cite{MagicAnimate} and (3) AnimateAnyone~\cite{AnimateAnyone} are characterized by their intricate appearance and temporal modelling, resulting in impressive performance. For quantitative comparison, we also compare with (4) DreamPose~\cite{DreamPose} which designs an adapter to incorporate features from human image. (5) DisCo~\cite{DISCO} is also an advanced diffusion-based method incorporating distinct conditioning modules for various elements such as pose, the foreground human subject, and the background. (6) BDMM~\cite{BDMM} shows good performance on fashion video generation. Due to the adoption of the same benchmarks and data splits, quantitative comparisons are conducted based on statistics directly cited from the original papers.

\begin{figure*}[t!]
  \centering
\includegraphics[width=\textwidth]{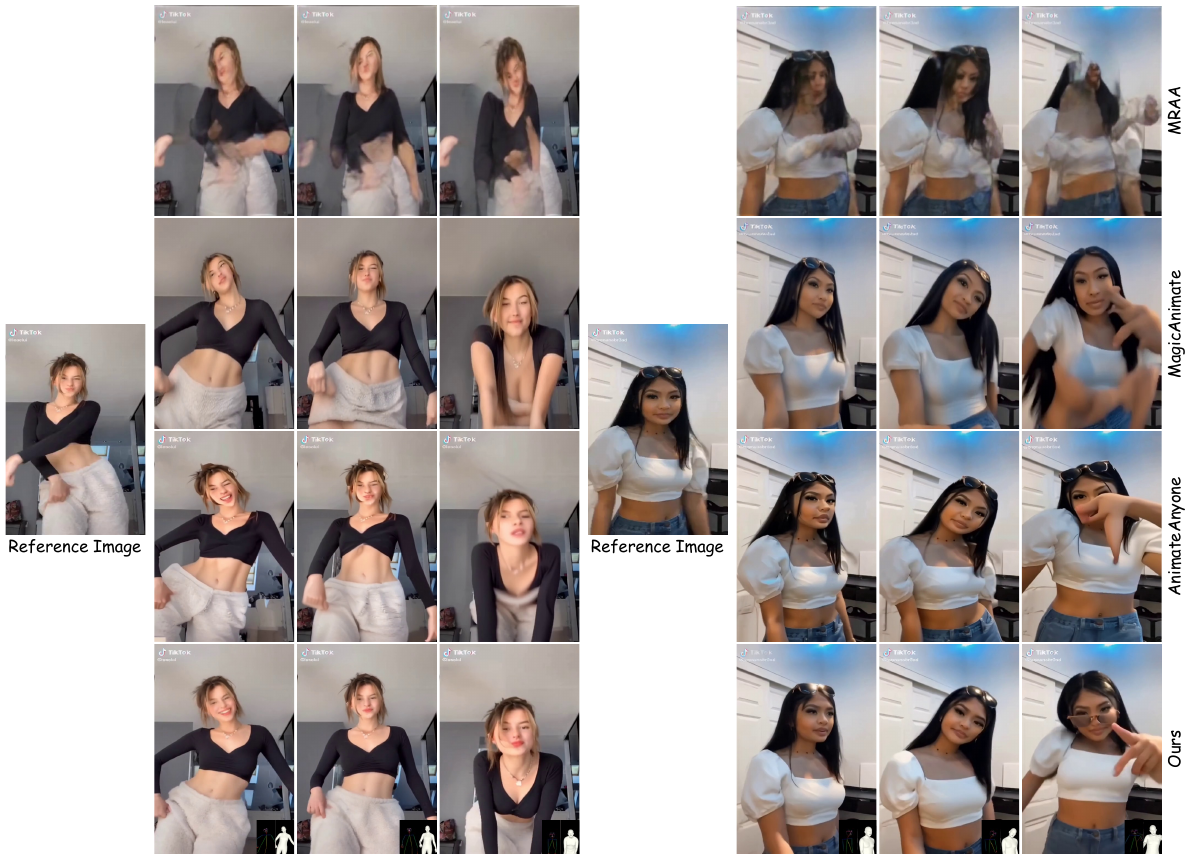}
   \caption{Qualitative comparison on TikTok dataset.} 
    \label{fig_comp_tiktok}
\end{figure*}

\subsection{Comparisons}
\textbf{Qualitative Comparison.}
From Fig.~\ref{fig_comp_tiktok}, it is evident that our method maintains better facial identity consistency during intense movements and successfully achieves complex hand gestures. An interesting case (fourth row, last column) shows that after vigorous motion, the sunglasses worn by the character generated by our method fall from the head and land perfectly on the eyes. This indicates that our method has a superior understanding of the physical world, further emphasizing its advantage in creating realistic animations. In summary, VividPose demonstrates clear advantages over state-of-the-art methods in several key areas. Our approach excels in preserving clothing texture details, maintaining high fidelity to facial identity, and accurately capturing body pose variations. Additionally, VividPose effectively handles complex scenarios involving hand crossing and dynamic movements. These strengths highlight the robustness and precision of our method, ensuring realistic and coherent human image animations that surpass the performance of existing techniques.

\textbf{Quantitative Comparison.}
We conducted extensive quantitative comparisons on the TikTok and UBCFashion datasets to evaluate the performance of VividPose against several state-of-the-art methods, including BDMM, DisCo, MagicAnimate, AnimateAnyone and DreamPose. The results on the TikTok dataset (Tab.~\ref{tab:quant_comp_tiktok}) show that VividPose outperforms other methods across multiple metrics, achieving the lowest FID score, the highest SSIM and PSNR, and one of the lowest LPIPS scores. These results indicate superior visual quality, structural similarity, and perceptual similarity. In terms of video fidelity, VividPose excels with the lowest FID-VID and FVD, demonstrating excellent temporal consistency. Performance further improves when VividPose is trained on our curated dataset. The results on the UBCFashion dataset (Tab.~\ref{tab:quant_comp_fashion}) highlight VividPose's strengths, achieving the highest SSIM and PSNR, and the lowest LPIPS, indicating its superior capability in preserving image quality and perceptual similarity. Additionally, VividPose achieves the lowest FVD, confirming its ability to generate temporally consistent videos. These results underscore VividPose's advantages in preserving fine-grained appearance details, maintaining facial identity, and accurately capturing complex body poses and movements. VividPose consistently outperforms existing methods in both image quality and video fidelity, making it a robust and reliable solution for human image animation.

\begin{figure*}[t]
  \centering
\includegraphics[width=\textwidth]{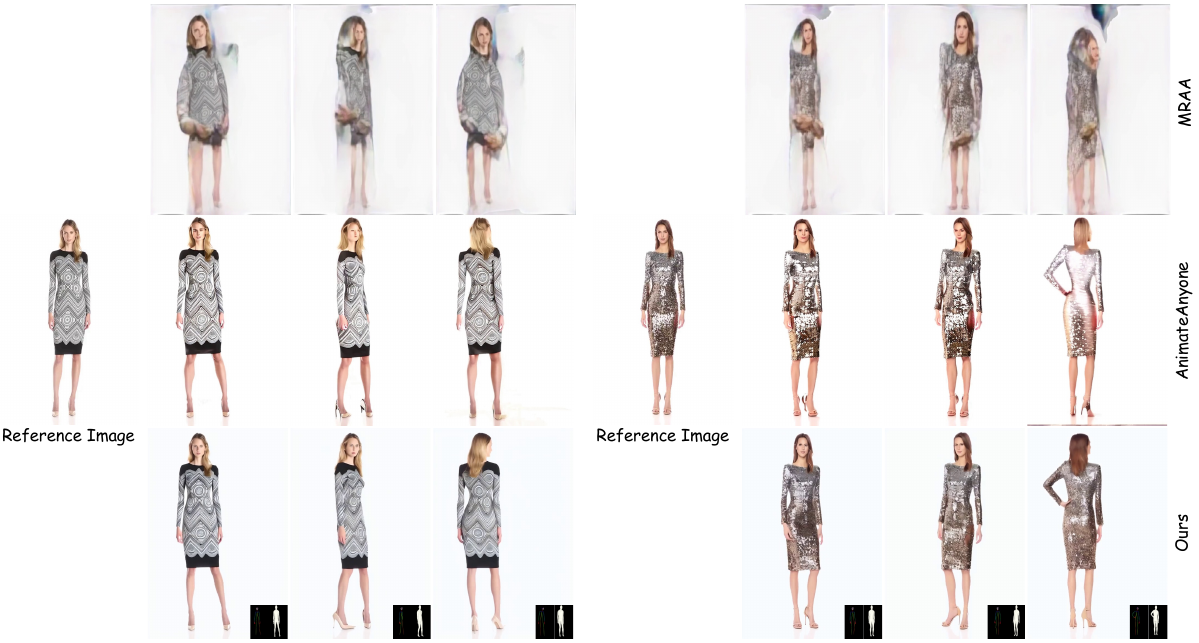}
   \caption{Qualitative comparison on UBCFashion dataset.} 
    \label{fig_comp_fashion}
\end{figure*}



\begin{figure*}[t]
  \centering
  \begin{minipage}{0.48\textwidth}
    \includegraphics[width=\linewidth]{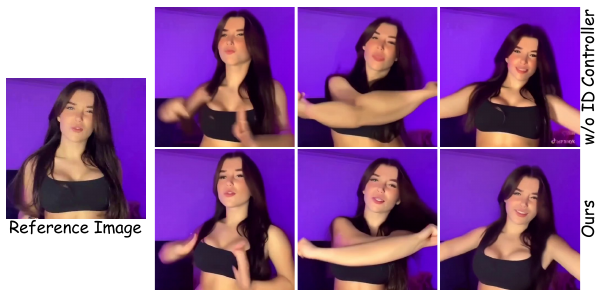}
    \caption{Qualitative ablation of ID controller.}
    \label{fig_ablation_id}
  \end{minipage}\hfill
  \begin{minipage}{0.48\textwidth}
    \includegraphics[width=\linewidth]{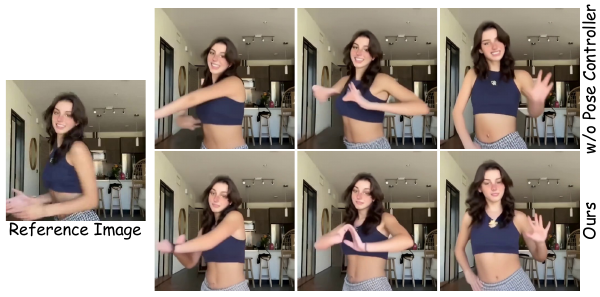}
    \caption{Qualitative ablation of Pose Controller.}
    \label{fig_ablation_pose}
  \end{minipage}
\end{figure*}

\begin{table*}[t!]
\centering
\caption{Quantitative comparisons on TikTok dataset of VividPose with several SOTA methods such as MRAA~\cite{MRAA}, DISCO~\cite{DISCO}, MagicAnimate~\cite{MagicAnimate} and AnimateAnyone~\cite{AnimateAnyone}.
$\downarrow$ indicates that the lower the better, and vice versa. Methods with $*$ directly use driving video frames as the input. $\dagger$ represents that VividPose is trained on our curated datasets. \label{tab:quant_comp_tiktok} }
\vspace{0.02in}
\begin{adjustbox}{width=\linewidth}
\begin{tabular}{lccccccc}
\toprule
\multirow{2}[2]{*}{Method}  & \multicolumn{5}{c}{\textbf{Image}} & \multicolumn{2}{c}{\textbf{Video}} \\ \cmidrule(lr){2-6} \cmidrule(lr){7-8}
 & \textbf{FID}\ $\downarrow$  & \textbf{SSIM}\ $\uparrow$ & \textbf{PSNR}\ $\uparrow$ & \textbf{LPIPS}\ $\downarrow$ & \textbf{L1}\ $\downarrow$ & \textbf{FID-VID}\  $\downarrow$ & \textbf{FVD}\ $\downarrow$ \\

\midrule
MRAA$^{*}$~\cite{MRAA}  &54.47  &0.672  & 29.39  &0.296  &3.21E-04  &66.36 &284.82\\
DisCo~\cite{DISCO} &50.68 &0.648  &28.81  &0.309  &4.27E-04  &69.68 &292.80  \\
MagicAnimate~\cite{MagicAnimate} & 32.09    &0.714  &29.16  &\textbf{0.239}  &3.13E-04  &21.75 &179.07  \\
AnimateAnyone~\cite{AnimateAnyone} & -    &0.718  &29.56  &0.285   &-  &- &171.9  \\

\midrule
VividPose  &31.89 &0.758 &29.83  & 0.261  & 6.89E-05 & 18.81
 & 152.97  \\
VividPose$^\dagger$  &\textbf{30.69}   &\textbf{0.774} &\textbf{30.07}  & 0.258  &\textbf{6.79E-05} & \textbf{15.12} & \textbf{139.54} \\
\bottomrule
\end{tabular}
\vspace{-5pt}
\end{adjustbox}
\end{table*}

\begin{table*}[t!]
\centering
\caption{Quantitative comparisons on UBCFashion dataset of VividPose with several SOTA methods, including MRAA~\cite{MRAA}, BDMM~\cite{BDMM}, DreamPose~\cite{DreamPose}, and AnimateAnyone~\cite{AnimateAnyone}.
$\downarrow$ indicates that the lower the better, and vice versa. $*$ denotes the result without sample finetuning.}
\label{tab:quant_comp_fashion}
\begin{adjustbox}{width=0.65\linewidth}
\begin{tabular}{lcccc}
\toprule
Method  & \textbf{SSIM}\ $\uparrow$ & \textbf{PSNR}\ $\uparrow$ & \textbf{LPIPS}\ $\downarrow$ & \textbf{FVD}\ $\downarrow$ \\

\midrule
MRAA~\cite{MRAA}  &0.749  &-  &0.212  &253.6 \\
BDMM~\cite{BDMM} &0.918 &24.07 &0.048 &148.3 \\
DreamPose~\cite{DreamPose} & 0.885    &-  &0.068  &238.7 \\
DreamPose$^*$ & 0.879 &34.75  &0.111  &279.6 \\
AnimateAnyone~\cite{AnimateAnyone} & 0.931    &38.49  &0.044  &81.6 \\

\midrule
VividPose  &\textbf{0.946}   &\textbf{41.56} &\textbf{0.0392}  & \textbf{62.3} \\
\bottomrule
\end{tabular}
\vspace{-5pt}
\end{adjustbox}
\end{table*}

\begin{table*}[t!]
\centering
\caption{Quantitative ablation study on TikTok dataset of VividPose with proposed modules. \textbf{ID} represents Identity-aware Appearance Controller. \textbf{Skeleton} denotes Pose Controller with Skeleton Maps. \textbf{Rendering} stands for Pose Controller with Rendering Maps.
\label{tab:ablation}}
\begin{adjustbox}{width=\linewidth}
\begin{tabular}{ccc cccccccc}
\toprule
  \multirow{2}[2]{*}{ID}  &  \multirow{2}[2]{*}{Skeleton} &
  \multirow{2}[2]{*}{Rendering} & \multicolumn{5}{c}{\textbf{Image}} & \multicolumn{2}{c}{\textbf{Video}} \\ \cmidrule(lr){4-8} \cmidrule(lr){9-10}&&&
  \textbf{FID}\ $\downarrow$  & \textbf{SSIM}\ $\uparrow$ & \textbf{PSNR}\ $\uparrow$ & \textbf{LPIPS}\ $\downarrow$ & \textbf{L1} $\downarrow$\ & \textbf{FID-VID}\  $\downarrow$ & \textbf{FVD}\  $\downarrow$ \\
\midrule
\xmark & \cmark & \cmark & 33.58 & 0.731 & 29.42  & 0.283 & 7.72E-05 & 19.93 & 191.75 \\
\cmark & \xmark & \cmark & 33.23 & 0.746 & 29.59  & 0.269  & 7.51E-05 & 19.56 & 175.03 \\
\cmark & \cmark & \xmark & 32.97 & 0.742 & 29.65  & 0.273 & 7.35E-05 & 19.42 & 166.64 \\
\cmark & \cmark & \cmark &\textbf{31.89} &\textbf{0.758} &\textbf{29.83}  & \textbf{0.261}  & \textbf{6.89E-05} & \textbf{18.81}
 & \textbf{152.97} \\
\bottomrule
\end{tabular}
\end{adjustbox}
\vspace{-10pt}
\end{table*}

\subsection{Ablation Study}
\textbf{Effectiveness of ID Controller.}
The quantitative and qualitative ablation study on the TikTok dataset, shown in Tab.~\ref{tab:ablation} and Fig.~\ref{fig_ablation_id}, underscores the effectiveness of the ID controller. Quantitatively, the inclusion of the ID controller improves image quality metrics: FID decreases from 33.58 to 31.89, SSIM increases from 0.731 to 0.758, and LPIPS decreases from 0.283 to 0.261. Qualitatively, the ID controller ensures better facial identity consistency, preserving key facial features across different poses. This is crucial for maintaining the recognizability of the subject, especially in dynamic scenarios. The ID controller performs well by effectively encoding and preserving identity-specific features, ensuring that the generated videos are both visually appealing and temporally coherent.

\textbf{Effectiveness of Pose Controller.}
The Pose Controller, which leverages both skeleton and rendering maps, is crucial for maintaining motion continuity and handling complex hand movements. As shown in Tab.~\ref{tab:ablation} and Fig.~\ref{fig_ablation_pose}, enabling both components results in better video quality metrics. Skeleton maps provide a sparse framework for overall body structure and motion, while rendering maps supplement detailed information about body shape and surface deformations. This dual approach allows for precise modeling of dynamic poses and complex hand movements, which are often challenging to replicate. These improvements highlight the pose controller's effectiveness in ensuring smooth and coherent motion, particularly in scenarios involving intricate hand movements and dynamic poses.
\section{Conclusion}
In this paper, we introduce VividPose, a novel end-to-end pipeline for human image animation based on Stable Video Diffusion (SVD). VividPose consists of two main modules: (1) The identity-aware appearance controller enhances human identity retention by incorporating facial information without compromising other appearance details. (2) The geometry-aware pose controller uses dense rendering maps from SMPL-X and sparse skeleton maps to accommodate diverse body shapes and hand effects. Extensive experiments on UBCFashion and TikTok benchmarks show that VividPose achieves state-of-the-art results in temporal consistency, visual fidelity, and generalization capabilities. It also performs well on our in-the-wild dataset, demonstrating robustness in real-world scenarios.



\newpage
\bibliographystyle{splncs04}
\bibliography{main}

\newpage
\appendix
\section{Appendix}
\subsection{More Details of Preliminary}
\subsubsection{SMPL-X \label{smplx}}

SMPL-X~\cite{SMPLX} is a 3D parametric human model that extends the capabilities of the SMPL~\cite{SMPL} model by incorporating detailed features of the face, hands, and body. SMPL-X represents the human body as a mesh with a fixed topology, controlled by shape, pose, and expression parameters. The shape parameters \(\beta\) capture identity-specific attributes, the pose parameters \(\theta\) encode joint rotations, and the expression parameters \(\psi\) model facial expressions. The model is defined as follows:
\begin{equation}
    M(\beta, \theta, \psi) = W(T(\beta, \theta, \psi), J(\beta), \theta, \mathcal{W})
\end{equation}
where \(M(\beta, \theta, \psi)\) represents the deformed 3D human model, \(T(\beta, \theta, \psi)\) is the template mesh, \(J(\beta)\) represents the joint locations, and \(\mathcal{W}\) are the blend weights. The function \(W(\cdot)\) applies the pose and shape blend weights to the template mesh, deforming it to match the desired pose and shape.

To obtain a 2D rendering map $I_{render}$ from the SMPL-X model, we use a differentiable rendering process. Given the camera projection function \(\Pi\), the rendering process can be formulated as:
\begin{equation}
    I_{render} = \Pi(M(\beta, \theta, \psi))
\end{equation}
This process ensures that the 3D human model is accurately projected onto the 2D image plane. For human image animation, the shape parameters \(\beta\) and expression parameters \(\psi\) are derived from the reference image, while the pose parameters \(\theta\) are obtained from the driving videos. We use the resulting rendering maps as pose conditions. These rendering maps ensure that the generated video's body shape accurately matches the reference image while following the dynamic poses from the driving videos, leading to realistic and coherent animations.

\subsubsection{SVD \label{appendix:svd}}
Stable Video Diffusion~\cite{SVD} (SVD) is a cutting-edge video generation model. This model builds upon latent diffusion model initially developed for 2D image synthesis, extending their capabilities to generate high-resolution, temporally consistent videos from text and image inputs. The original UNet processes image data with the input shape \(\mathbb{R}^{B \times C \times H \times W}\), where \(B\) is the batch size, \(C\) is the number of channels, \(H\) is the height, and \(W\) is the width. For video data, the input shape is \(\mathbb{R}^{B \times T \times C \times H \times W}\), where \(T\) represents the video length. To adapt video data to the existing image model's structure, the first two dimensions are merged, resulting in \(\mathbb{R}^{(B \times T) \times C \times H \times W}\).

SVD introduces two types of temporal layers: 3D convolution layers and temporal attention layers. For the 3D convolution layers, the data is reshaped from \(\mathbb{R}^{(B \times T) \times C \times H \times W}\) to \(\mathbb{R}^{B \times C \times T \times H \times W}\) before applying the convolution, and then reshaped back after the operation. For the temporal self-attention layers, the data is reshaped from \(\mathbb{R}^{(B \times T) \times C \times H \times W}\) to \(\mathbb{R}^{(B \times H \times W) \times T \times C}\), ensuring that the temporal dynamics are effectively captured. Temporal layers are also integrated into the VAE decoder. As for training, a significant improvement is the transition from the DDPM~\cite{DDPM} noise scheduler to the EDM~\cite{EDM} scheduler, along with adopting EDM's sampling method. Traditional DDPM models use discrete timesteps \(t\) for denoising, but EDM introduces the continuous noise scale \(\sigma_t\). By inputting \(\sigma_t\) into the model, it enables more flexible and effective sampling. During the sampling process, continuous noise strengths are used instead of discrete timesteps. The end-to-end training paradigm and pipeline of SVD demonstrate strong advantages in maintaining temporal consistency in video generation, making it particularly suitable for human image animation.







\newpage

\end{document}